\documentclass{article}
\usepackage{spconf,amsmath,graphicx, algorithm, algorithmic}
\usepackage{xcolor}
\usepackage{multirow}
\usepackage{flushend}

\usepackage{enumitem}
\setlist{nosep, leftmargin=14pt}

\usepackage{mwe} 

\newcommand{\etal}{\textit{et al.}}

\title{Smart Split-Federated Learning Over Noisy Channels for Embryo Image Segmentation}
\name{Zahra Hafezi Kafshgari, Ivan V. Baji\'{c}, and Parvaneh Saeedi\thanks{This work was supported in part by the NSERC grants RGPIN-2018-05164, RGPIN-2021-02485, and RGPAS-2021-00038.}
}
\address{School of Engineering Science, Simon Fraser University, Burnaby, BC, Canada \\
** Accepted at ICASSP 2023. The final version appears in the conference proceedings. ** }
\begin{document}
%
\maketitle
\begin{abstract}
Split-Federated (SplitFed) learning is an extension of federated learning that places minimal requirements on the clients' computing infrastructure, since only a small portion of the overall model is deployed on the clients' hardware. In SplitFed learning, feature values, gradient updates, and model updates are transferred across communication channels. In this paper, we study the effects of noise in the communication channels on the learning process and the quality of the final model. We propose a smart averaging strategy for SplitFed learning with the goal of improving resilience against channel noise. Experiments on a segmentation model for embryo images shows that the proposed smart averaging strategy is able to tolerate two orders of magnitude stronger noise in the communication channels compared to conventional averaging, while still maintaining the accuracy of the final model.     

\end{abstract}
\section{Introduction}
\label{sec:intro}
\vspace{-5pt}
Federated learning (FL)~\cite{mcmahan2017communication} enables training of machine learning models without sharing data. This is important in certain scenarios, including medical data analysis, where data cannot be shared due to privacy or other regulations. However, FL may still be challenging for healthcare applications, because it assumes that participating clients -- medical clinics in this case -- have the computing infrastructure to train large machine learning models. 
During FL, each client is responsible for training a local model, which is then sent to a central server where local models are aggregated into a global model~\cite{yang2019federated,li2020federated}. 
The reality is that many medical clinics do not have the computing infrastructure for training local models, which are of the same size as the global model. 

On the other hand, split learning (SL)~\cite{SL} is a learning approach capable of addressing computational power imbalance between clients' devices and servers. In SL, the front-end of the model, which is usually a small part of the overall model, is located at the client's device, while the larger back-end is located at the server. During training, features (also known as ``smashed data'') are sent from the front-end to the back-end, while gradient updates are sent from the back-end to the front-end. SL can be used in a federated scenario across multiple clients~\cite{splitAVG,SFLenergy}, in which case it is sometimes referred to as split-federated (SplitFed) learning~\cite{splitfed}. 
SplitFed learning allows both privacy control and computational balancing between the clients and the server
~\cite{comparison_FL_SL_SFL, SFL_healthcare}.
However, it requires more communication than FL, because not only local models, but also features and gradient updates in each training batch need to be exchanged between the clients and the server.

No communication channel is perfect. Therefore, the data exchanged between the clients and the server will be noisy, and this will impact the SplitFed learning process. Specifically, noise in the features and gradient updates during local model training will degrade the quality of local models, whereas noise in the model transfer will impact the global model. To our knowledge, these issues were not previously studied in the context of SplitFed learning. In this paper, we examine the impact of communication noise on SplitFed learning of an embryo image segmentation model, and propose an averaging strategy -- termed \emph{Smart SplitFed} -- that minimizes its effects. The paper is organized as follows. In Section~\ref{sec:related}, we present a brief overview of related work. The proposed method is described in Section~\ref{sec:proposed}. Experiments are presented in Section~\ref{sec:experiments} followed by conclusions in Section~\ref{sec:conclusions}. 
\vspace{-5pt}
\section{Related works}
\label{sec:related}
\vspace{-5pt}
The key step in FL is aggregation (averaging) of local models into a global model. The simplest way of doing this is to take arithmetic average of local models' weights to update global model weights~\cite{mcmahan2017communication}. However, this simple strategy is agnostic to client heterogeneity, for example, different distribution of data across clients. Hence, several approaches were developed to address this issue. 

In~\cite{shi2021fed}, an approach called Fed-ensemble was introduced to average \textit{K} randomly trained models in the federation step to improve the generalization of the global model. In~\cite{robust_FL}, the authors study robust heterogeneous FL including clients with mislabeled data. Note that in~\cite{robust_FL}, data labels are considered to be subject to noise, whereas communication links are assumed to be perfect. Hence,~\cite{robust_FL} is complementary to the scenario studied here. 
Along similar lines,~\cite{auto_robust_FL} created corrupted clients by randomly shuffling labels and adding Gaussian noise to input data, then minimized a weighted sum of empirical risks to alleviate the effects of corruption. 
It was shown that global performance drops drastically as more clients become corrupted.
Yang \etal~\cite{FL_noisyLabels} also considered noisy labels in FL and proposed sharing centriods of local feature vectors for model aggregation and label correction. 
In~\cite{li2020federated}, an averaging method called FedProx was proposed to deal with device and data heterogeneity in FL. None of these approaches considered communication noise. 

FL was combined with split learning to create SplitFed in~\cite{splitfed}, to address both privacy issues and computational imbalance between the clients and the server. 
Two versions of SplitFed were proposed: vanilla SplitFed, in which the labels are shared with the server while the input data remains at the client; and U-shaped SplitFed (Fig.~\ref{fig:SplitFed_U-Net}), in which both input data and labels remain at the client. 
Simple averaging was used for both versions of SplitFed in~\cite{splitfed}, which, as our results will show, is very sensitive to communication noise.

SplitFed is gaining popularity in healthcare applications. Poirot \etal~\cite{SFL_healthcare} utilized it in experiments on fundus photos and chest X-rays. Gawali \etal~\cite{comparison_FL_SL_SFL} reviewed FL, SL, and SplitFed approaches in healthcare, and introduced another version of SplitFed called SpltFedv3, in which labels are shared, but client models are not averaged. Instead, only server-side model weights are averaged. 
In~\cite{splitAVG}, SplitAVG was proposed for the purpose of averaging features sent to the server from heterogeneous clients. 
Both vanilla and U-shaped SpliFed were examined in~\cite{splitAVG}. 
In~\cite{SFLenergy}, heterogeneity in terms of energy harvesting capabilities of clients was tackled. However, none of these earlier works considered communication noise in SplitFed learning, which is our main focus. 




\vspace{-5pt}
\section{Proposed Method}
\label{sec:proposed}
\vspace{-5pt}
\subsection{Segmentation model}
\begin{figure}[t]
    \centering
    \includegraphics[height=5.8 cm, width=9 cm]{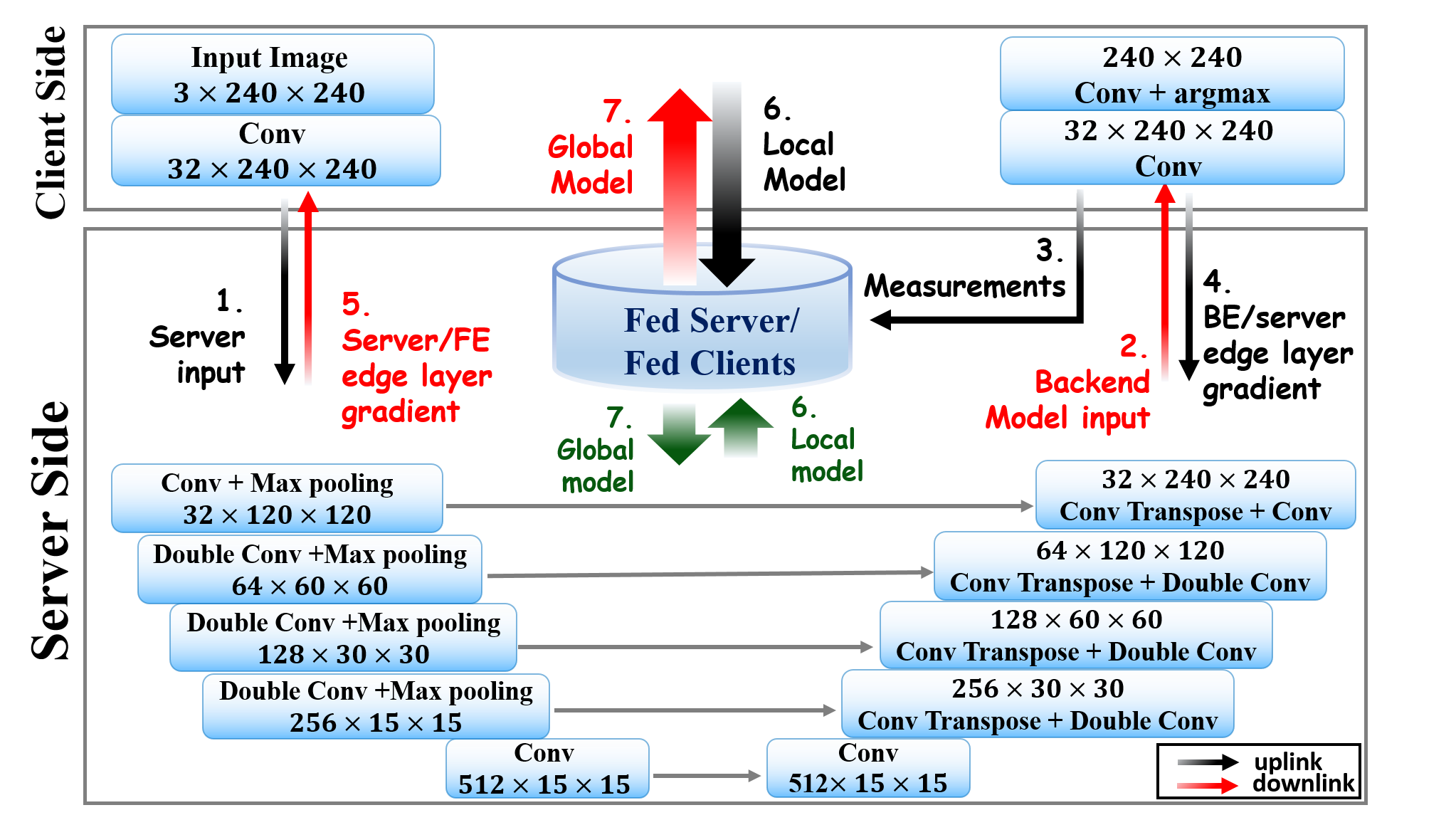}
    \caption{SplitFed U-Net used in our work.} 
    \label{fig:SplitFed_U-Net}
\end{figure}

We study training of a SplitFed U-Net, shown in Fig.~\ref{fig:SplitFed_U-Net}, for segmenting embryo images. The overall U-Net consists of five downsampling blocks between the input and the bottleneck, and five upsampling blocks between the bottleneck and the output. Each block contains two convolutional layers with the kernel size of $3\times3$, followed by batch normalization and \texttt{ReLU} activation. Each downsampling block ends with a $2\times2$ max-pooling layer; the number of filters in downsampling blocks is 32, 64, 128, 256, and 512, respectively. Each upsampling block starts with a $2\times2$ upsampling layer; the number of filters in upsampling blocks is 512, 256, 128, 64, and 32, respectively. The last convolutional layer is followed by an output \texttt{argmax} layer. The model is split at the first and last convolutional layer, as shown in the figure, so that the first colvolutional layer (front-end, FE) and the last convolutional layer together with the output \texttt{argmax} layer (back-end, BE) are on the client side, while the rest of the model is on the server. Links from the client to the server (uplinks) are shown in black, while the links from the server to the client (downlinks) are shown in red. 

\subsection{Training procedure}
The system adopts sequential training where the server trains with each of the $N$ clients in turn. The server runs several \textit{local epochs} of training with a specific client. At the end of this process, both the client and the server store their weights from the best local epoch, i.e., the local epoch with the lowest validation loss. Let $W^{\text{S}}_i$ be the server weights from the best local epoch while training with client $i$, and $W^{\text{C}}_i=\{W^{\text{FE}}_i, W^{\text{BE}}_i\}$ be the weights of client $i$, which include both FE and BE weights, from the same epoch. One pass across all clients is called a \textit{global epoch}. At the end of each global epoch, $W^{\text{S}}_i$ and $W^{\text{C}}_i$ are averaged into $\overline{W}^\text{S}$ and $\overline{W}^\text{C}$, respectively, to create a global model $W=\{\overline{W}^\text{S}, \overline{W}^\text{C}\}$ before the next global epoch starts. Averaging is discussed in Section~\ref{sec:averaging} below.

Within each local epoch, in the forward pass, features produced by the FE are sent to the server, then processed by the server-side part of the model, then features from the server are sent to the client-side BE. In the backpropagation pass, gradient updates from the BE are sent to the server, then back-propagated through the server-side model, then sent to the FE. 
Data that is exchanged between the server and the client -- features in the forward pass and gradients in backpropagation -- are subject to noise during transmission.


\subsection{Averaging}
\label{sec:averaging}
Two popular model averaging methods in federated learning are naive averaging and federated averaging~\cite{mcmahan2017communication}. In naive averaging, each client receives the same weight, so with $N$ clients the averaging would look like:
\begin{equation}
    \overline{W}^{\text{S}} = \frac{1}{N}\sum_{i=1}^N W^{\text{S}}_i, \qquad
    \overline{W}^{\text{C}} =
    \frac{1}{N}\sum_{i=1}^N W^{\text{C}}_i.
    \label{eq:naive_SplitFed}
\end{equation}
In federated averaging, the weights are proportional to the number of data samples at each client. If $m_i$ is the number of data samples at client $i$, federated averaging would be: 
\begin{equation}
    \overline{W}^{\text{S}} = \frac{1}{\sum_{i}m_i}\sum_{i=1}^N m_iW^{\text{S}}_i, \quad
    \overline{W}^{\text{C}} = \frac{1}{\sum_{i}m_i}
    \sum_{i=1}^N m_iW^{\text{C}}_i.
    \label{eq:SplitFed_AVG}
\end{equation}

Neither of these popular averaging strategies takes into account the quality of clients' models. However, clients with noisy links 
tend to train less accurately than those with clean links, as our results will show. Hence, a ``smart'' averaging strategy should take that into account. Here we propose one such strategy whose goal is to take both potential data imbalance (different $m_i$'s) and client's 
loss (which is impacted by communication noise) in the averaging process.

After completing local epochs with client $i$ and selecting the model from the best epoch, it is possible to compute the loss on each training sample $j$ on that client, $\mathcal{L}^j_i$. 
Hence, the following statistics can be computed:
\begin{equation}
    \mu_i = \texttt{mean}\{\mathcal{L}^1_i,...,\mathcal{L}^{m_i}_i\}, \quad \sigma_i = \texttt{std}\{\mathcal{L}^1_i,...,\mathcal{L}^{m_i}_i\}.
\end{equation}
We want to assign lower weight to the clients with higher loss. However, the average loss estimate $\mu_i$ based on a relatively small $m_i$ is unreliable. Hence, to reflect this uncertainty, an upper bound of a confidence interval for the mean loss can be taken as an indicator for the (un)reliability of that client. If $\mathcal{L}^j_i$'s were normally distributed, the upper bound of the 95\% confidence interval for the mean loss would be 
\begin{equation}
    b_i = \mu_i + 2\sigma_i.
\end{equation}
This is what we take as the indicator of (un)reliability of client $i$. All clients' indicators are collected in the vector $\mathbf{b}=[b_1,...,b_N]^{\top}$, then the vector of quality scores $\mathbf{q}=[q_1,...,q_N]^{\top}$ is computed as 
\begin{equation}
    \mathbf{q} = \texttt{softmax}[\alpha\cdot(1-\mathbf{b})],
    \label{eq:quality_scores}
\end{equation}
where $\alpha$ is a hyperparameter, set to $\alpha=10$ in our experiments. The key point about~(\ref{eq:quality_scores}) is that (un)reliability indicators in $\mathbf{b}$ appear with the negative sign, so that less reliable clients (higher $b_i$) receive lower quality scores (lower $q_i$). \texttt{softmax} serves to emphasize the more reliable clients.

Besides quality scores, we also take the data distribution across clients into account, in the same way as federated averaging does. Specifically, let $\mathbf{d}=[m_1,...,m_N]^{\top}/(\sum_i{m_i})$ be the vector containing relative amounts of data at each client. Our final weights $\mathbf{r}=[r_1,...,r_N]^{\top}$ are obtained by combining quality scores and data distribution as follows:
\begin{equation}
    \mathbf{r}=\frac{\mathbf{q}\odot\mathbf{d}}{\mathbf{q}^{\top}\mathbf{d}},
\end{equation}
where $\odot$ is the Hadamard 
product. Note that $\sum_{i}r_i = 1$. Finally, \textit{Smart SplitFed} averaging is performed as:
\begin{equation}
    \overline{W}^{\text{S}} = \sum_{i=1}^N r_iW^{\text{S}}_i, \quad
    \overline{W}^{\text{C}} = 
    \sum_{i=1}^N r_iW^{\text{C}}_i.
    \label{eq:smart_splitfed}
\end{equation}

\section{Experiments}
\label{sec:experiments}
\vspace{-5pt}
\subsection{Experimental setup}
\vspace{-5pt}
We train the SplitFed U-Net model (Fig.~\ref{fig:SplitFed_U-Net}) on multi-label segmentation of human blastocyst components \cite{blastnet,blastosys_data}. 
There are 815 embryo images in total, each with a segmentation mask that includes five components: Background (BG), Zona Pellucida (ZP), Trophectoderm (TE), Blastocoel (BL), and Inner Cell Mass (ICM).
A test set of 100 images is saved for testing the global model. There are $N=5$ clients and the data is non-uniformly distributed across clients, 
with 210, 120, 85, 180 and 120 samples respectively on clients 1, 2,..., 5. Each client uses 85\% of its data for training and 15\% for validation. Clients gets trained over 12 local and 10 global epochs.

Input images are resized to $240\times240$, and augmentation is performed using horizontal and vertical flipping, and rotation of up to $\pm35^{\circ}$. 
Dice coefficient~\cite{dice_loss} is chosen as the loss function, and the system is trained with the Adam optimizer with the initial learning rate set to $10^{-3}$. 
Inspired by Softcast~\cite{softcast}, channel noise is simulated by adding white Gaussian noise with zero mean and standard deviation of $\sigma_{\text{noise}}$ to all transmitted and received data of clients 3, 4, and 5. This means that noise is added to features transmitted in the forward pass, gradient updates in the backwards pass, as well as $b_i$ and $W_i^{\text{C}}$ (sent from client $i$ to the server after local epochs) and $\overline{W}^{\text{C}}$ (sent from server to all clients after averaging, but contaminated only for noisy clients). Noise is added starting at global epoch 5 for client 3, global epoch 4 for client 4, and global epoch 3 for client 5.  

\begin{table}[tb]
\caption{\small{A comparison of Naive SplitFed, SplitFedAVG (AVG), and Smart SplitFed on the common test set at different values of $\sigma_{\text{noise}}$. Acc is the overall accuracy; the last five columns show IOU of Background (\textbf{BG}), Zona Pellucida (\textbf{ZP}), Trophectoderm (\textbf{TE}), Inner Cell Mass (\textbf{ICM}), and Blastocoel (\textbf{BL}). Loss being `nan' indicates that the method failed to converge.}} 
\label{tab:results}
\center
\setlength{\tabcolsep}{3.5pt}
\renewcommand{\arraystretch}{1}
\footnotesize
\begin{tabular}{|c|c|c|c|c|c|c|c|c|}
\hline 

\( \sigma_{\text{noise}} \) &  \textbf{ Method}& \textbf{Loss} &  \textbf{Acc}  &  \textbf{BG} & \textbf{ZP} &  \textbf{TE} & \textbf{ICM} & \textbf{BL}   \\ \hline \hline

\multirow{3}{*}{$0$}  
 & Naive & 0.08 & 93.39 & 0.93 & 0.78 & 0.74 & 0.85 & 0.88  \\ 
 & AVG & 0.10 & 92.50 & 0.92 &  0.75 &  0.74 &  0.83 & 0.87  \\ 
 & Smart  &  \textbf{0.06}  & \textbf{ 93.60 }  & \textbf{0.94}  & \textbf{0.80 } & \textbf{ 0.75}   & \textbf{0.85}  & \textbf{0.88}   \\
 \hline \hline

\multirow{3}{*}{$2\cdot10^{-4}$} 
 &  Naive &   0.17  &  89.00 &   0.90  &  0.71 &  0.70 & 0.73 & 0.79    \\ 
 &  AVG &   0.13  &  89.44 &   0.88  & 0.71 &  0.71 & 0.77 & 0.83    \\
 & Smart  & \textbf{0.07} &  \textbf{93.52}  & \textbf{0.95} & \textbf{0.78} & \textbf{0.73}  &  \textbf{0.83}  & \textbf{0.88} \\
 \hline \hline

\multirow{3}{*}{$6\cdot10^{-4}$} 
 &  Naive &   0.21 &  82.02 &   0.74 &  0.51 &  0.65 &  0.75 & 0.79 \\ 
 &  AVG &    0.16 &  87.10 &    0.86 &   0.64  &  0.65 &   0.74 & 0.79  \\ 
 & Smart  &  \textbf{0.07} &   \textbf{93.18}   &   0.94 &    \textbf{0.77}  &   \textbf{0.72} &   \textbf{0.83}  & \textbf{0.87} \\ 
\hline \hline

\multirow{3}{*}{$10^{-3}$} 
 &  Naive &    nan &  40.27  &    0.40 &   0.00  &  0.00 &   0.00 & 0.00   \\
 &  AVG &    0.23 &  76.49 &    0.59 &   0.50  &  0.67 &   0.73 & 0.78   \\
 &  Smart  &  \textbf{0.12} &  \textbf{92.97} &  \textbf{0.93} &  \textbf{0.78} &  \textbf{0.74} &  \textbf{0.83} & \textbf{0.87} \\
 \hline \hline

\multirow{3}{*}{$10^{-2}$} 
 &  Naive &   nan  & 40.27   &   0.40 &  0.00  & 0.00  &  0.00 &  0.00  \\
 &  AVG &   nan  & 40.27  &   0.40 &  0.00  & 0.00  &  0.00 &  0.00  \\ 
 & Smart  &    \textbf{0.08} &  \textbf{93.05}   &   \textbf{0.94} &   \textbf{0.79} &  \textbf{0.74} &  \textbf{0.85}  & \textbf{0.88 } \\ 
\hline \hline

\multirow{3}{*}{$10^{-1}$} 
&  Naive &  nan &  40.27  &  0.40 &   0.00 & 0.00  &  0.00 & 0.00   \\ 
&  AVG &   {nan}  & 40.27  &   0.40 &  0.00  & 0.00  &  0.00 &  0.00  \\
&  Smart  &   \textbf{0.08}  &  \textbf{92.66}  &  \textbf{0.93} &  \textbf{0.78} &  \textbf{0.74} &  \textbf{0.85} & \textbf{0.86} \\ 
\hline \hline

\multirow{3}{*}{$5\cdot10^{-1}$} 
 &  Naive &  nan &  40.27  &  0.40 &  0.00 & 0.00  & 0.00 & 0.00   \\
 &  AVG &   {nan}  & 40.27  &   0.40 &  0.00  & 0.00  &  0.00 &  0.00  \\ 
 &  Smart  &   \textbf{0.06} &  \textbf{93.12} &  \textbf{0.94} &  \textbf{0.78} &  \textbf{0.73} &  \textbf{0.82} & \textbf{0.87} \\ 
\hline  

\end{tabular}
\label{tab1}

\end{table}

\vspace{-10pt}
\subsection{Results}
\vspace{-5pt}
We compare the Naive SplitFed~(\ref{eq:naive_SplitFed}), SplitFedAVG~(\ref{eq:SplitFed_AVG}) and Smart SplitFed~(\ref{eq:smart_splitfed}) in terms of the global model accuracy and loss value at the end of the 10$^{\text{th}}$ global epoch. The results are shown in Table~\ref{tab:results} in terms of the Intersection Over Union (IOU) of the five embryo segments and the overall accuracy, defined as the percentage of correctly classified pixels. The best results are shown in bold. Experiments were conducted for seven values of $\sigma_{\text{noise}}$ shown in the table. In cases where the loss is shown as `nan', the training did not converge, likely due to gradient noise contamination during backpropagation. This happens to Naive SplitFed and SplitFedAVG with stronger noise. In such cases, the model defaults to predicting only the background, and the IOU of other components is zero.  

With $\sigma_{\text{noise}}=0$, all averaging methods perform comparably, and similar to BLAST-Net~\cite{blastnet}. However, as the noise increases, the advantage of Smart SplitFed over the other two methods becomes obvious. Naive SplitFed fails to converge already at $\sigma_{\text{noise}}=10^{-3}$, while SplitFedAVG fails to converge starting with $\sigma_{\text{noise}}=10^{-2}$. Smart SplitFed, on the other hand, is very robust to noise and barely loses any accuracy all the way up to $\sigma_{\text{noise}}=5\cdot10^{-1}$. 

\begin{figure}[t]
    \centering
    \includegraphics[ width=8.5 cm]{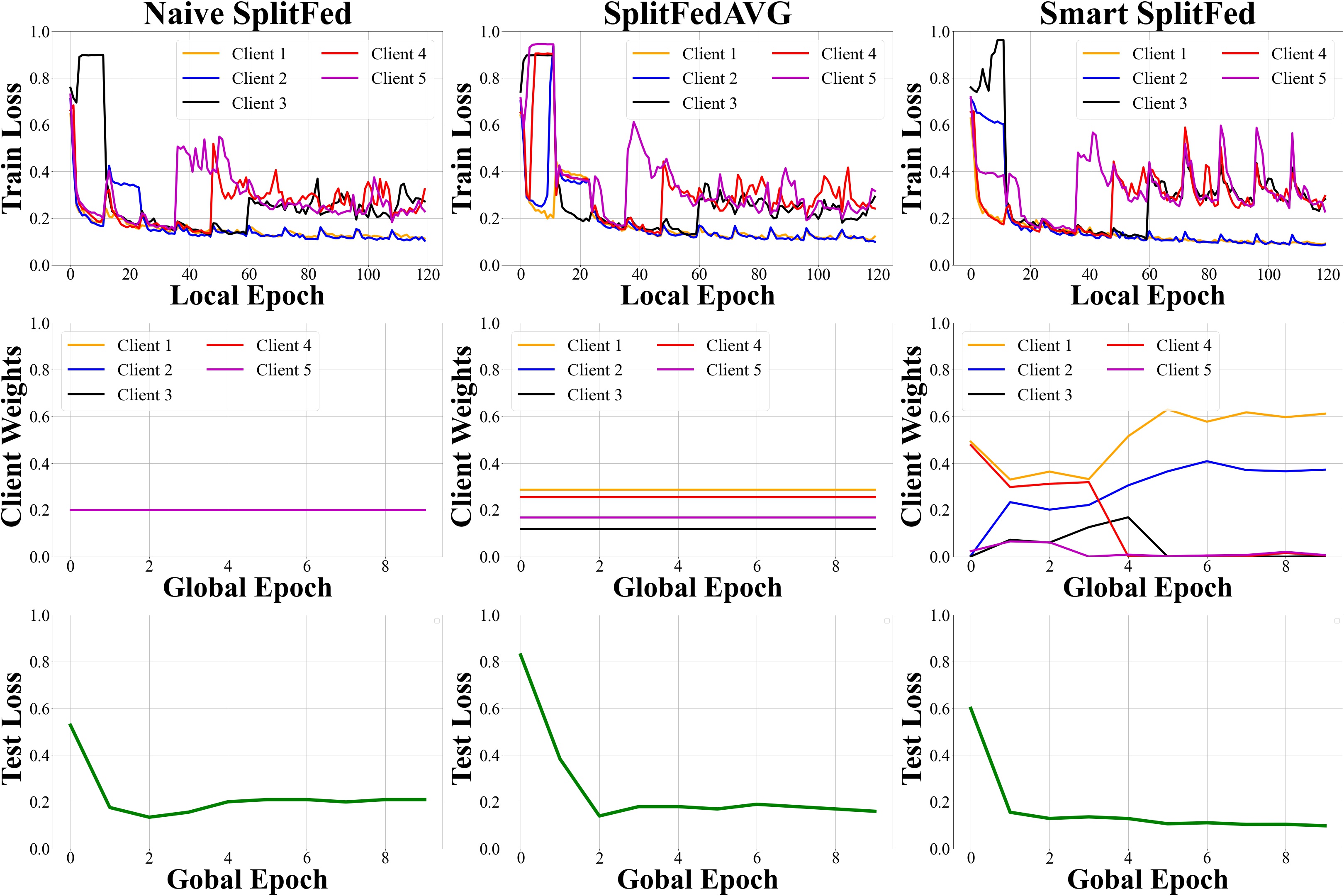}
    \caption{A comparison of Naive SplitFed, SplitFedAVG, and Smart SplitFed at  $\sigma_{\text{noise}} = 6\cdot 10^{-4}$. Top row: training loss at each client. Middle row: clients' weights used in averaging. Bottom row: test loss at each global epoch.} 
    \label{fig:loss_vs_epoch}
\end{figure}

\begin{figure}[!htb]
    \centering
    \includegraphics[ width=8.5 cm]{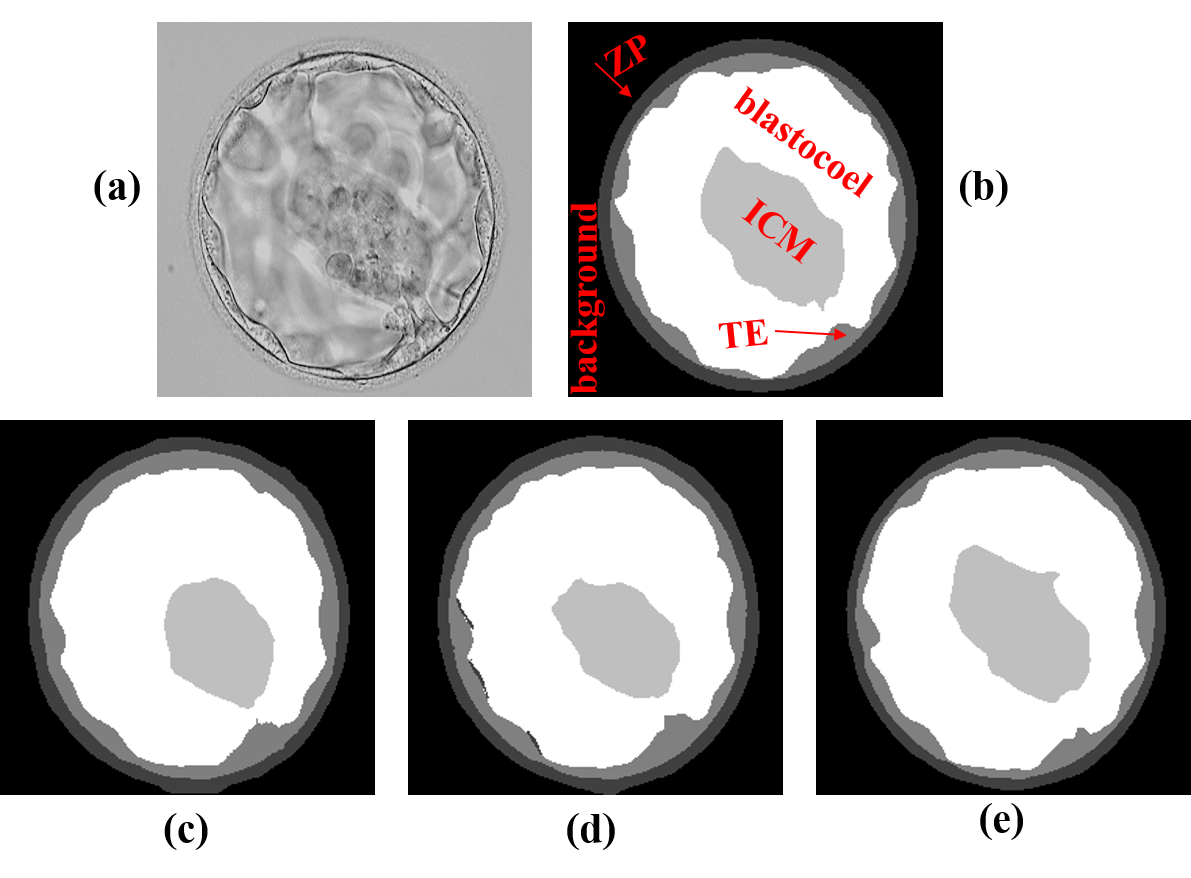}
    \caption{(a) Input image, (b) ground truth segmentation, and predictions made by models trained at $\sigma_{\text{noise}} = 6\cdot 10^{-4}$ using (c) Naive SplitFed, (d) SplitFedAVG, and (e) Smart SplitFed. 
    } 
    \label{fig:predicted_segmentation_maps}
\end{figure}


  



To probe further, Fig.~\ref{fig:loss_vs_epoch} shows the evolution of training at  $\sigma_{\text{noise}}=6\cdot10^{-4}$, because this is the highest noise level in the experiments where all averaging strategies converged. Noise is introduced at global epochs 3, 4, and 5 (i.e., after local epochs 36, 48, and 60) at clients 5, 4, and 3. We can observe the corresponding increase in the local training loss at those clients in the top-row graphs in Fig.~\ref{fig:loss_vs_epoch}. The middle-row graphs show the clients' weights used in averaging. Naive SplitFed uses equal weights, whereas SplitFedAVG uses weights proportional to the amount of data at each client. Neither of the two methods adjusts its weights during training. However, Smart SplitFed adjusts to the increased loss of clients 3, 4, and 5 by reducing their weights. Indeed, after global epoch 5, only clients 1 and 2 (with clean communication links) receive weights that are significantly different from zero. The result is the lower test loss at the end of training, as shown in the bottom-row graphs, as well as Table~\ref{tab:results}. 

Finally, in Fig.~\ref{fig:predicted_segmentation_maps} we visualize some of the segmentation masks. The top row shows an input image and the corresponding ground-truth segmentation mask with the five segmentation components indicated on the mask. The bottom row shows the predicted segmentation masks using models trained at $\sigma_{\text{noise}}=6\cdot10^{-4}$ by Naive SplitFed, SplitFedAVG, and Smart SplitFed. We note that the model trained using Smart SplitFed has produces the most accurate prediction. In particular, its prediction about the size and shape of the ICM is clearly better than those produced by the models trained with Naive SplitFed and SplitFedAVG.


\section{Conclusions}
\label{sec:conclusions}
The focus of this paper was split-federated (SplitFed) learning in the presence of noise on the links between clients and the server. Traditional averaging methods were not designed for such a scenario, and our experiments on embryo image segmentation showed they are fairly fragile in this context. We proposed a novel averaging method called \textit{Smart SplitFed}, which suppresses the effects of communication noise by monitoring the local loss values at each client and adjusting the averaging weights accordingly. The proposed method was shown to be much more robust to noise on the communication links compared to conventional methods and lead to more accurate global models.  


\small
\bibliographystyle{IEEEbib}
\bibliography{refs}

\end{document}